\documentclass[letterpaper]{article} 
\usepackage{aaai24}  
\usepackage{times}  
\usepackage{helvet}  
\usepackage{courier}  
\usepackage[hyphens]{url}  
\usepackage{graphicx} 
\usepackage{natbib}  
\usepackage{caption} 
\usepackage{algorithm}
\usepackage{algorithmic}
\usepackage{algorithm}
\usepackage{algorithmic}
\usepackage{times}
\usepackage{epsfig}
\usepackage{graphicx}
\usepackage{amsmath}
\usepackage{amssymb}
\usepackage{multirow}
\usepackage{multicol}
\usepackage{adjustbox}
\usepackage{booktabs}
\usepackage{makecell}
\usepackage{bm}
\usepackage{float}
\usepackage{caption}
\usepackage{latexsym}
\usepackage{subcaption}
\usepackage{fancyhdr}
\usepackage{cancel}
\input{def.set}
\usepackage[dvipsnames]{xcolor}
\usepackage[accsupp]{axessibility}

\usepackage{bibentry}
\def\semichecked{\checkmark\!\!\!\raisebox{0.4 em}{\tiny$\smallsetminus$}}

\usepackage{newfloat}
\usepackage{listings}
\DeclareCaptionStyle{ruled}{labelfont=normalfont,labelsep=colon,strut=off} 
\floatstyle{ruled}
\newfloat{listing}{tb}{lst}{}
\floatname{listing}{Listing}

\title{TD$^2$-Net: Toward Denoising and Debiasing for Dynamic Scene Graph Generation}
\author{
    Xin Lin\textsuperscript{\rm 1}{\thanks{Corresponding Author}},
    Chong Shi\textsuperscript{\rm 1},
    Yibing Zhan\textsuperscript{\rm 2},
    Zuopeng Yang\textsuperscript{\rm 1}\textsuperscript{\thefootnote},
    Yaqi Wu\textsuperscript{\rm 1},
    Dacheng Tao\textsuperscript{\rm 3}
}

\affiliations{
    \textsuperscript{\rm 1} Guangzhou University \\
    \textsuperscript{\rm 2} JD Explore Academy \\
    \textsuperscript{\rm 3} The University of Sydney \\

    linxin94@gzhu.edu.cn, \{shichong, winnerwu\}@e.gzhu.edu.cn, zybjy@mail.ustc.edu.cn\\
    \{yzpeng44, dacheng.tao\}@gmail.com

}

\begin{document}

\maketitle

\begin{abstract}
Dynamic scene graph generation (SGG) focuses on detecting objects in a video and determining their pairwise relationships. Existing dynamic SGG methods usually suffer from several issues, including 1) Contextual noise, as some frames might contain occluded and blurred objects. 2) Label bias, primarily due to the high imbalance between a few positive relationship samples and numerous negative ones. Additionally, the distribution of relationships exhibits a long-tailed pattern. To address the above problems, in this paper, we introduce a network named TD$^2$-Net that aims at denoising and debiasing for dynamic SGG. Specifically, we first propose a denoising spatio-temporal transformer module that enhances object representation with robust contextual information. This is achieved by designing a differentiable Top-K object selector that utilizes the gumbel-softmax sampling strategy to select the relevant neighborhood for each object.
Second, we introduce an asymmetrical reweighting loss to relieve the issue of label bias. This loss function integrates asymmetry focusing factors and the volume of samples to adjust the weights assigned to individual samples. Systematic experimental results demonstrate the superiority of our proposed TD$^2$-Net over existing state-of-the-art approaches on Action Genome databases. In more detail, TD$^2$-Net outperforms the second-best competitors by 12.7 \% on mean-Recall@10 for predicate classification. 

\end{abstract}

\section{Introduction}

A scene graph is a graph-structured representation that uses nodes to represent objects and edges to represent relationships in an image. It provides a practical approach for scene understanding, serving as a bridge between visual and language modalities, and is widely applied in various fields such as image captioning~\cite{yang2020hierarchical}, image retrieval ~\cite{johnson2015image}, and visual question answering ~\cite{yang2020hierarchical}. While the development of scene graph generation (SGG) of images has been satisfactory, research on dynamic scene graph generation of videos is still in its infancy.

Dynamic SGG aims to detect the objects for each frame and the relationships among them. Such a detailed and structured video understanding is akin to how humans perceive real-world activities. Existing approaches ~\cite{cong2021spatial, nag2023unbiased, feng2023exploiting} in dynamic SGG predominantly utilize transformers~\cite{vaswani2017attention} for context modeling to acquire spatial-temporal information of objects or predicates. Furthermore, some methods employ strategies such as temporal prior inference~\cite{wang2022dynamic}, uncertainty-aware learning, and memory-guided training~\cite{nag2023unbiased} to achieve unbiased dynamic SGG.

\begin{figure}[t]
\begin{center}
    \centering
    \includegraphics[scale=0.4]{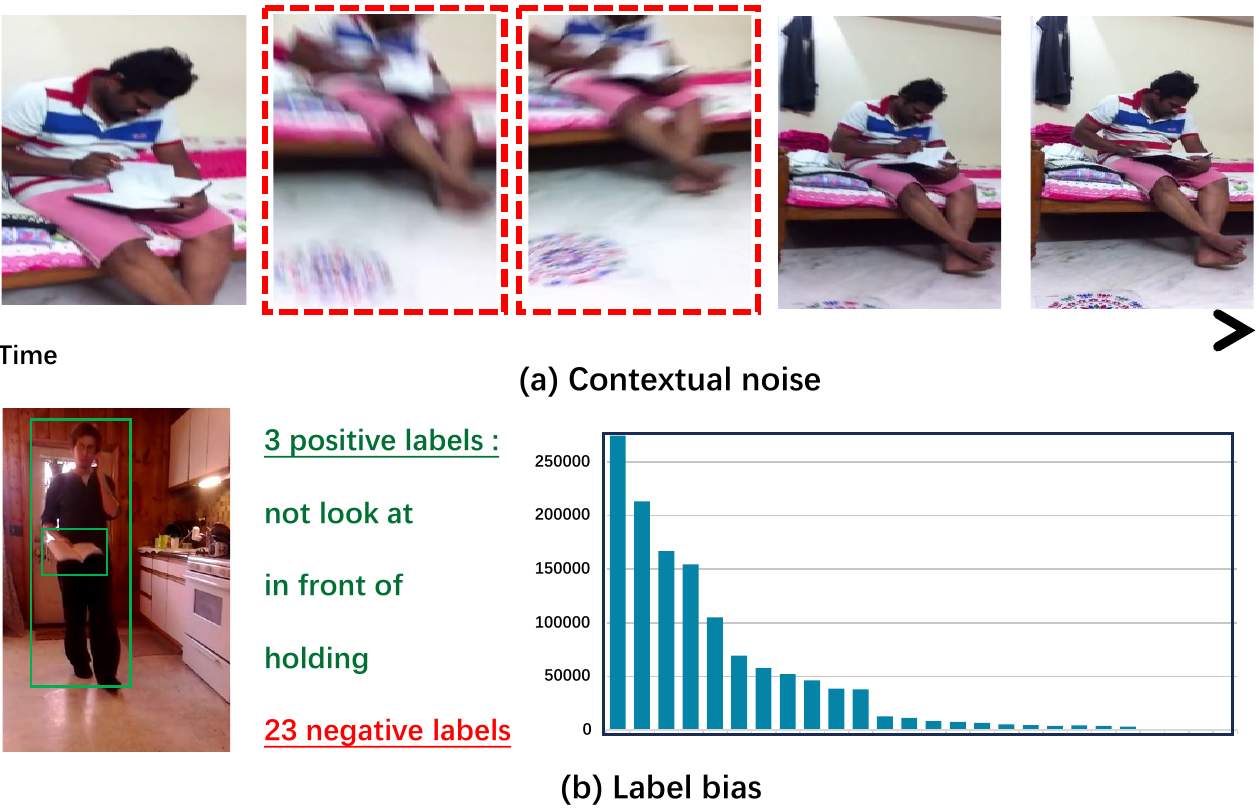}
\vspace{-6mm}
    \captionof 
    {figure}{ (a) Contextual noise. A significant proportion of objects may be occluded or affected by camera motion blur.
    (b) Label bias. As shown in the left example of two objects, the quantity of positive relationship labels is significantly less than that of negative ones, causing a negative-positive imbalance. Furthermore, as shown in the right tabular, the distribution of relationships exhibits a long-tailed trend.
    }\label{intro1}
\end{center}
\end{figure}
However, as shown in Figure~\ref{intro1}, the current dynamic SGG still faces two main problems. Firstly, there is the issue of contextual noise, as videos consist of noisy and correlated sequences of frames~\cite{buch2022revisiting}. A significant fraction of objects may be occluded or affected by camera motion blur, as depicted in Figure~\ref{intro1}(a). When acquiring contextual information for objects or relationships, the irrelevant objects may introduce redundant or erroneous information, leading to unnecessary computational overhead and reducing the accuracy of the model's predictions. Secondly, there is the problem of label bias. As depicted in Figure~\ref{intro1}(b), the predicate classes in standard datasets exhibit two types of imbalances, including positive-negative imbalance and head-tail imbalance. The former may result in the potential underestimation of the gradient of positive labels during training, as negative labels are often much more numerous than positive labels. The long-tailed distribution of positive labels may lead to the model's inability to recognize rare positive samples effectively, thereby reducing the accuracy and diversity of the model's predictions.

To address the abovementioned issues, we propose a network named (TD$^2$-Net) that aims at denoising and debiasing for dynamic SGG. Firstly, we introduce the denoising spatio-temporal transformer (D-Trans) module to tackle the contextual noise. Specifically, we achieve preliminary object matching by considering objects' appearance and spatial location consistency across frames. Furthermore, to eliminate noisy spatio-temporal information caused by irrelevant objects, we introduce the Gumbel-Softmax sampling strategy~\cite{jang2016categorical}. This ensures that each object only aligns with the most relevant neighboring objects within a video, selecting more appropriate contextual information. Afterward, we enhance the object representations by aggregating contextual information that have been selected.

Secondly, we introduce the asymmetrical reweighting loss (AR-Loss) to address the label bias issue in relationship prediction. To tackle the positive-negative imbalance problem, we utilize different values of focusing factors for positive and negative samples, controlling their contribution to the loss function. Specifically, we set the focusing factor of positive samples to be higher than that of negative ones, leading the model to place more emphasis on positive samples. Moreover, we incorporate the concept of the effective number of samples, as described by~\cite{cui2019class}, to mitigate the problem of head-tail imbalance. This adjustment allows the model to learn meaningful features from positive samples, despite their rarity.

In summary, the contributions of this study are two-fold: (1) D-Trans for enhancing object features with denoising contextual information; (2) AR-Loss to deal with both positive-negative imbalance and head-tail imbalance in relationship prediction. The efficacy of the proposed TD$^2$-Net is systematically evaluated on the video scene graph generation benchmark dataset. Experimental results show that our TD$^2$-Net consistently outperforms state-of-the-art methods.

\section{Related Work}

\subsection{Image Scene Graph Generation}
Existing works in image SGG (ImgSGG) typically focus on context modeling or tackling the class imbalance problem (\textit{i.e.}, the long-tailed distribution). Several context modeling strategies~\cite{lin2020gps,li2021bipartite}, have been proposed to learn discriminative object representation by exploring various message passing mechanisms. Specifically, Lin \textit{et al.} build a heterophily-aware message-passing scheme to distinguish the heterophily and homophily between objects/relationships ~\cite{lin2022hl}. Tang \textit{et al.} proposes a dynamic tree structure to capture task-specific visual contexts ~\cite{tang2019learning}. To handle the class imbalance issue, Lin \textit{et al.} propose a group diversity enhancement module to relieve low diversity in relationship predictions ~\cite{lin2022ru}. Zheng \textit{et al.} relieves the ambiguous entity-predicate matching caused by the predicate’s semantic overlap by prototype regularization ~\cite{zheng2023prototype}. However, modeling each frame in long videos can result in high computational complexity and redundant information. Current ImgSGG methods prioritize spatial over temporal information, thus missing inter-frame correlations. Furthermore, unlike ImgSGG, which focuses on a single-label biased problem, video SGG or dynamic SGG deals with a more complicated multi-label biased problem.

\subsection{Video Scene Graph Generation}

Compared with ImgSGG, video SGG (VidSGG) or dynamic SGG is more challenging because it needs to consider the spatio-temporal context in adjacent frames. Existing VidSGG methods can be roughly divided into two categories, including tracklet-based and frame-based approaches. Specifically, for the tracklet-based method, each graph node is an object tracklet within a video segment~\cite{shang2017video,shang2019annotating} or the entire video~\cite{liu2020beyond,gao2022classification,gao2023compositional}. 
For the frame-based method, each graph node is an object box, similar to ImgSGG, but the visual relation triplets are dynamic throughout the entire video sequence ~\cite{ji2020action,feng2023exploiting,cong2021spatial,li2022dynamic}. Additionally, researchers have begun to focus on addressing the issue of biased prediction in VidSGG. In more detail, Wang \textit{et al.} utilizes temporal prior knowledge as an inductive bias to generate dynamic scene graph~\cite{wang2022dynamic}. Nag \textit{et al.} learns to synthesize unbiased relationship representations using memory-guided training and attenuates the predictive uncertainty of visual relations using a Gaussian Mixture Model~\cite{nag2023unbiased}. However, current VidSGG methods still face two main problems. Firstly, there is the issue of contextual noise caused by certain frames in a video that might contain occluded or blurred objects. Secondly, the predicate classes exhibit two types of imbalances, including positive-negative imbalance and head-tail imbalance.

\begin{figure*}[ht!]
\centering
\includegraphics[width=1.\linewidth]{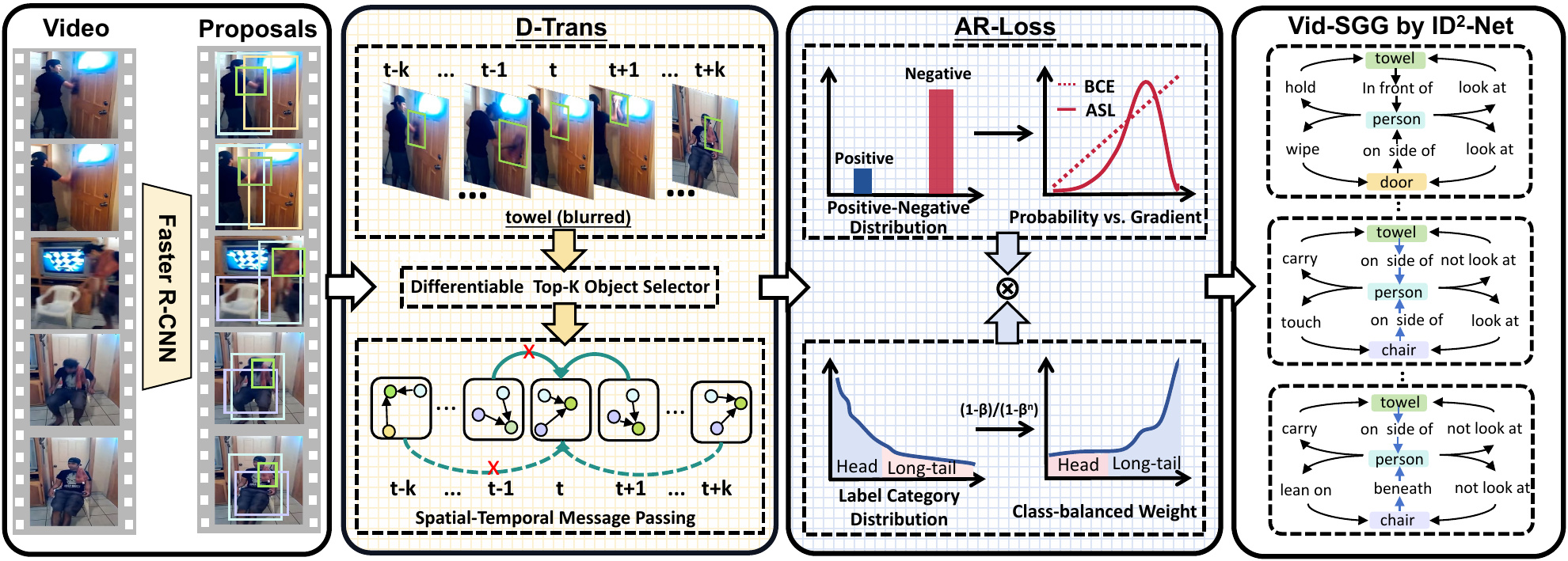}
\caption{The framework of TD$^2$-Net. TD$^2$-Net adopts Faster-RCNN to generate initial object proposals for each RGB frame in a video. It includes two new modules for dynamic scene graph generation: (1) a novel transformer module named D-Trans that enhances object feature with robust contextual information (2) a new loss function named AR-Loss that takes into account both positive-negative imbalance and head-tail imbalance in relationship prediction.}
\label{fig:model1}
\end{figure*}  

\section{Method}
The framework of the proposed TD$^2$-Net is illustrated in Figure~\ref{fig:model1}. We employ Faster R-CNN \cite{ren2015faster} to obtain object proposals for each video frame. We adopt the same way as~\cite{cong2021spatial} to obtain the feature for each proposal. The appearance features, spatial feature, and classification score vector for the $i$-th object is denoted as $\bx_i$, $\bb_{i}$, and $\bp_{i}$, respectively.
We further extract features from the union box of one pair of proposal $i$ and $j$, denoted as $x_{ij}$. To achieve denoising and debiasing for dynamic SGG, we have made two contributions in this work. Firstly, a denoising spatio-temporal transformer (D-Trans) module is introduced to address the issue of context noise. Secondly, an asymmetrical reweighting loss (AR-Loss) is introduced to tackle the label bias problem. In the below, we will describe these two components sequentially.

\subsection{Denosing Spatio-Temporal Transformer}

Existing works on VidSGG~\cite{gao2021video,nag2023unbiased,feng2023exploiting} typically gather contextual information by tracking or matching objects in sequential frames. However, these methods are prone to being affected by contextual noise, leading to some spurious and redundant object correlations, which reduces the accuracy of model predictions and results in unnecessary resource consumption. To alleviate this issue, we propose the D-Trans module, which includes a differentiable Top-$\rm K$ object selector and spatial-temporal message passing.

\subsubsection{Differentiable Top-$\rm K$ Object Selector.}
Inspired by the video graph representation method proposed in~\cite{xiao2022video}, we employ a score function based on objects' appearance and spatial location to ensure temporal consistency across frames. Specifically, the matching score for two objects $i$ and $j$ in different frames can be expressed as follows:

\begin{equation}\label{score_1}
    g_{ij}=   \psi\left(\bp_{i}, \bp_{j}\right)+\operatorname{IoU}\left(\bb_{i}, 
 \bb_{j}\right),
\end{equation}
where $\psi$ and IoU denote the cosine similarity and intersection-over-union functions, respectively. Detected objects in consecutive frames are linked to the target objects by greedily maximizing Eq.~\eqref{score_1} frame by frame. By aligning objects within a video, we ensure the consistency of the object representations for the graphs constructed at different frames. By gathering all the appearance features of the aligned objects for the $i$-th object, we can obtain its neighborhood feature matrix $\bZ_i$.

Compared with the object matching strategy in existing frame-based dynamic SGG works, Eq.~\eqref{score_1} has two advantages. First, compared with utilizing the predicted object label~\cite{nag2023unbiased}, it utilizes the score vector of objects which can better reflect the uncertainty of the prediction. Besides, the IoU function can ensure temporal consistency. Second, compared with utilizing Hungarian matching~\cite{feng2023exploiting}, it enjoys a low computational complexity.

However, as the number of video frames increases in long videos, the number of objects associated with a target object also increases. In order to dynamically eliminate unreliable spatio-temporal information caused by irrelevant objects, inspired by the method ~\cite{gao2023mist} of using a segment selector to extract the relevant frames for the given question, we propose a differentiable Top-$\rm K$ selector to choose the most relevant contextual information from neighboring objects for the target object, which is defined as follows:

\begin{equation}\label{selector}
    \bF_i = \sigma(\underset{{\rm Top}\text{-}{\rm K}}{\rm selector} ({\rm softmax}( \frac{\bx_i \bK^T}{\sqrt{d_k} }, \bV   )  ) ),
\end{equation}
where $\sigma$ represents the function that flattens the matrix into a vector by row. $\bK\!\!=\!\!\bV\!\!=\!\!\bZ_i+\bE_i$, $d_k$ denotes the dimension of $\bK$, the $\bE_i$ represents positional encodings~~\cite{vaswani2017attention} that provide temporal location information for each aligned object relative to the $i$-th object. In more detail, the Top-$\rm K$ selection can be implemented by extending the gumbel-softmax trick~\cite{jang2016categorical} or based on optimal transport formulations ~\cite{xie2020differentiable} for ranking and sorting. In this paper, we conduct gumbel-softmax sampling $\rm K$ times with replacement to achieve Top-$\rm K$ selection. Note that we sample the objects with replacement, as certain objects may only exist in a few frames. In such cases, we aim to guide the model learns to select the most related object by re-sampling it instead of forcing it to select irrelevant objects, as sampling without replacement does.

\subsubsection{Spatial-Temporal Message Passing.}\label{sst}
Given the selected neighborhood for each object, we further stack $N$ standard multi-head attention (MHA) layers~\cite{vaswani2017attention} to enhance the object representations by aggregating information from other aligned objects from the adjacent frames within a video:

\begin{equation}\label{ntrans}
    \bX^{'}  = {\rm MHA}_{\rm{temporal}}^{(N)}(\Phi^t_q(\bX+\bE),\Phi^t_k(\bF),\Phi^t_v(\bF)) ,
\end{equation}

where $\Phi^t_q$, $\Phi^t_k$, and $\Phi^t_v$ are linear transformation. In addition, we apply $M$ MHA layers to reason over the object spatial interactions as follows:
\begin{equation}\label{ntrans_s}
    \hat\bX  = {\rm MHA}_{\rm{spatial}}^{(M)}(\Phi^s_q(\bX^{'}_f),\Phi^s_k(\bX^{'}_f),\Phi^s_v(\bX^{'}_f)) ,
\end{equation}
where $\Phi^s_q$, $\Phi^s_k$, and $\Phi^s_v$ denote linear transformation, $\bX^{'}_f$ denotes the representations of objects from the same frame.

In summary, our motivation for the D-Trans module is that it models the change of object behaviors and thus infers dynamic actions. Also, it is helpful in improving the objects’ appearance feature in cases where the object at certain frames suffers from motion blur or partial occlusion.

\subsection{Asymmetrical Reweighting Loss}
After obtaining the refined object representation by D-Trans, we further propose an asymmetrical reweighting loss (AR-Loss) to mitigate the issue of label bias in relationship prediction. We adopt the same approach as in STTran~\cite{cong2021spatial} to obtain the classification score vector of the relationship between two objects $i$ and $j$ as follows:
\begin{equation}\label{rel}
\bp_{ij}={\phi}({\rm RTrans}([\bW\!_s\hat\bx_i,\bW\!_o\hat\bx_j,\bx_{ij},\bc_i,\bc_j])),
\end{equation}
where $\rm RTrans$ denotes that we utilize the same structure as our proposed spatial-temporal message passing module to refine the relationship feature. $\phi$ indicates the classification operation. $\bc_i$, $\bc_j$ denote the semantic embedding of the $i$-th object and $j$-th object. $\bW\!_s$, $\bW\!_o$ are projection matrices for fusion, $\left[,\right]$ represents the concatenation operation.

Binary cross-entropy loss~\cite{nag2023unbiased,wang2022dynamic} and multi-label margin loss~\cite{cong2021spatial} have been widely adopted for optimization in previous dynamic SGG models. However, the aforementioned loss functions consider each sample to be equally significant and assign them the same weight, which may not be suitable to alleviate the issue of label bias. To address this problem, we first revisit the focal loss~\cite{lin2017focal}, which is a traditional solution to mitigate the positive-negative imbalance issue. It adjusts the loss contribution of easy and hard samples, which reduces the influence of the majority of negative samples:

\begin{equation}\label{fl}
    \mathcal{L}_{fl}\left(\bp_{ij}\right) = \begin{cases} \mathcal{L}_{fl}^{+} =  \left ( 1-\bp_{ij} \right )^{\gamma} {\rm log} \left ( \bp_{ij} \right ), 
  & \text{ if } y=1 \\
  \mathcal{L}_{fl}^{-} = \bp_{ij}^{\gamma} {\rm log}(1-\bp_{ij}),
  & \text{ if } y=0
\end{cases}
\end{equation}
where $y$ is the ground-truth label, and $ \mathcal{L}_{fl}^{+}$ and $\mathcal{L}_{fl}^{-}$ represent the loss function of positive and negative samples, respectively. $\gamma$ denotes the focusing parameter. By setting $\gamma > 0$, It reduces the impact of easy negatives on the loss function. However, the focal loss may not sufficiently address the following issues:

\begin{itemize}
    \item  \textbf{Positive-Negative Imbalance.} Most object pairs contain fewer positive labels and more negative labels on average. A High value of $\gamma$ in Eq.~\eqref{fl} sufficiently down-weights the contribution from easy negatives but may eliminate the gradients from the tail positive samples.  
     \item  \textbf{Head-Tail Imbalance.} Due to the long-tailed distribution of the datasets, the head-tail imbalance exists in the positive samples. This imbalance between different categories of positive samples may lead to the model failing to recognize rare positive samples. 
\end{itemize}

Thus, we decouple the focusing levels of the positive and negative samples to alleviate the positive-negative imbalance. Specifically, we set $\gamma^{+}$ and $\gamma^{-}$ to be the positive and negative focusing parameters, respectively. Furthermore, to mitigate the impact of head-tail imbalance in the positive samples, inspired by~\cite{cui2019class}, which adjusts sample weights using the effective number of samples for each class, we defined $\omega_{cb}$ as follows to adjust the weights assigned to individual sample:

\begin{equation}\label{cb}
  \omega_{cb} = \frac{1-\beta }{1-\beta^{n_{\hat{y}}} }, 
\end{equation}
where $n_{\hat{y}}$ is the number of samples of the ground-truth class $\hat{y}$ in the training set. A higher value of $\omega_{cb}$ for tail samples will increase their weight, encouraging the model to pay more attention to the positive tail samples and vice versa. The hyper-parameter $\beta \in \left[0,1\right)$ controls the rate at which the weight grows as $n_{\hat{y}}$ increases.

After applying the asymmetric focusing factors $\gamma^{+}$, $\gamma^{-}$, and the effective number of samples $\omega_{cb}$ into our AR-Loss, we obtain the loss function as follows: 
\begin{equation}\label{ar}
    \mathcal{L}_{ar}(\bp_{ij}) = \begin{cases} \calL_{ar}^{+} = \omega_{cb}   \left ( 1-\bp_{ij} \right )^{\gamma ^{+}}  {\rm log}\left ( \bp_{ij} \right ), 
  & \text{ if } y=1 \\
  \calL_{ar}^{-} = \bp_{ij}^{\gamma ^{-}}  {\rm log}(1-\bp_{ij}).
  & \text{ if } y=0
\end{cases}
\end{equation}
Note that $\gamma^{+}=\gamma^{-}=0$ and $\omega_{cb}$=1 yields binary cross-entropy. Since we are interested in emphasizing the contribution of positive samples, we set $\gamma^{-} \ge \gamma^{+}$. We achieve better control over the contribution of positive and negative samples through Eq.~\eqref{ar}, which assists the network in learning meaningful features from positive samples, despite their rarity. Thus, AR-loss can simultaneously address the positive-negative imbalance and head-tail imbalance.

\subsection{VidSGG by TD$^2$-Net}
During training, the overall loss function $\calL$ for TD$^2$-Net can be expressed as follows:
\begin{equation}\label{loss}
    \mathcal{L} = \mathcal{L}_{obj} + \mathcal{L}_{ar},
\end{equation}
where $\calL_{obj}$ denotes the cross entropy loss for object classification.

During testing, the score of each relationship triplet $<$subject-predicate-object$>$ is computed as:
\begin{equation}\label{predicate}
    s_{rel}= s_{sub} * s_p * s_{obj},
\end{equation}
where $s_{sub}$, $s_p$, $s_{obj}$ are the predicted score of subject, predicate, and object, respectively.

\begin{table*}[ht]
\centering
\setlength{\tabcolsep}{1.6mm}
\begin{tabular}{ccccccccccccc}
\toprule[2pt]
\midrule[1pt]
 \multirow{3}*{Method} &  \multicolumn{6}{c}{With Constraint} &\multicolumn{6}{c}{No Constraint} \cr
    \cmidrule(lr){2-7} \cmidrule(lr){8-13}  
  & \multicolumn{2}{c}{PredCLS} & \multicolumn{2}{c}{SGCLS} & \multicolumn{2}{c}{SGDET} & \multicolumn{2}{c}{PredCLS} & \multicolumn{2}{c}{SGCLS} & \multicolumn{2}{c}{SGDET} \cr
    \cmidrule(lr){2-3} \cmidrule(lr){4-5} \cmidrule(lr){6-7} \cmidrule(lr){8-9} \cmidrule(lr){10-11} \cmidrule(lr){12-13} 
  &R@10  &R@50 &R@10&R@50 &R@10  &R@50 &R@10   &R@50 &R@10  &R@50 &R@10&R@50\\
  \hline \hline
  VRD~\shortcite{lu2016visual}& 51.7& 54.7& 32.4& 33.3& 19.2& 26.0& 59.6& 99.2& 39.2& 52.6& 19.1& 40.5\\
  Motif Freq\shortcite{zellers2018neural}& 62.4& 65.1& 40.8& 41.9& 23.7& 33.3& 73.4& 99.6& 50.4& 64.2& 22.8& 46.4\\
  MSDN~\shortcite{li2017scene} & 65.5& 68.5& 43.9& 45.1& 24.1& 34.5& 74.9& 99.0& 51.2& 65.0& 23.1& 46.5\\
  VCTREE~\shortcite{tang2019learning}& 66.0& 69.3& 44.1& 45.3& 24.4& 34.7& 75.5& 99.3& 52.4& 65.1& 23.9& 46.8\\
  RelDN~\shortcite{zhang2019graphical} & 66.3& 69.5& 44.3& 45.4& 24.5& 34.9& 75.7& 99.0& 52.9& 65.1& 24.1& 46.8\\
  GPS-Net~\shortcite{lin2020gps}& 66.8& 69.9& 45.3& 46.5& 24.7& 35.1& 76.0& 99.5& 53.6& 66.0& 24.4&47.3\\
  STTran~\shortcite{cong2021spatial}& 68.6& 71.8 & 46.4  & 47.5 & 25.2& 37.0 & 77.9  & 99.1& 54.0 & 66.4&24.6 &48.8 \\
  TPI~\shortcite{wang2022dynamic} &69.7&72.6 &47.2  &48.3  &26.2  &\bf{37.4}  &-  &- &-  &- &-  &-\\

  TEMP~\shortcite{nag2023unbiased} &68.8  &71.5 &47.2  &48.3 &28.1 &34.9 &80.4  &99.4 &56.3  &67.9 &29.8  &46.4\\\hline

  {\bf TD$^2$-Net (P)} &\bf{70.1}  &\bf{73.1}&\bf{51.1}&\bf{52.1}    &\bf{28.7}&{37.1}&\bf{81.7}&\bf{99.8} &\bf{57.2}&\bf{69.8}    &\bf{30.5}&\bf{49.3} \\
  {\bf TD$^2$-Net} &67.8  &70.8&49.1&50.2    &27.2&36.7     &78.2&99.2 &55.1&67.3    &28.1&48.4\\
\midrule[1pt]
\bottomrule[1pt]
\end{tabular}

\caption{Comparisons with state-of-the-art on the Action Genome dataset. The same object detector is used in all baselines for fair comparison. TD$^2$-Net (P) indicate that we set $\omega_{cb}$ as 1 in AR-Loss. The best methods are marked according to formats under each setting.} 

\label{tb1}
\end{table*}

\section{Experiments}
\subsection{Dataset and Evaluation Setting}\label{data_eval}
\noindent\textbf{Dataset.}
Our experiments are conducted on the AG dataset~\cite{ji2020action}, which is the benchmark dataset of dynamic scene graph generation. AG is built upon the Charades dataset~\cite{sigurdsson2016hollywood} and provides frame-level scene graph labels with a total of 234,253 frames in 9,848 video clips. In AG, there are 36 types of entities and 26 types of relations in the label annotations. Such 26 types of relations are divided into three classes (\textit{i.e.,} attention, spatial, and contacting relations). The attention relations are used to describe if a person is looking at an object or not. The spatial relations specify the relative position. The contacting relations represent different ways of contacting in particular.

\noindent\textbf{Evaluation Setting.}
We use the same data and evaluation metrics that have been widely adopted in recent works~\cite{ji2020action, cong2021spatial, nag2023unbiased, wang2022dynamic}. Specifically, We make the evaluation of TD$^2$-Net on the AG dataset under three conventional tasks below: (1) Predicate Classification (PredCLS): Given the ground-truth object bounding boxes and categories, the model needs to predict predicate categories; (2) Scene Graph Classification (SGCLS): Given the ground-truth bounding boxes of objects, the model needs to predict both the object and relationship categories; (3) Scene Graph Detection (SGDET): Given an image, the model detects object and predict relationship categories between each pair of objects. All algorithms are evaluated using the Recall@$K$ (R@K) and mean-Recall@$K$ (mR@K) metrics, for $K\text{=}[10,50]$. Evaluation is conducted under two setups: \textbf{With Constraints} and \textbf{No Constraints} to make a fair and sufficient comparison with baselines. In more detail, \textbf{With Constraints} is the most stringent since it only chooses one predicate for each entity pair. \textbf{No Constraints} allows multiple predictions of relations for each entity pair, taking top $100$ predicates for all pairs in a single frame. 

\begin{table*}[!htbp]
\centering
\setlength{\tabcolsep}{0.5mm}
\begin{tabular}{ccccccccccccccccccc}
\toprule[2pt]
\midrule[1pt]
 \multirow{3}*{Method} &  \multicolumn{6}{c}{With Constraint} &\multicolumn{6}{c}{No Constraint} \cr
    \cmidrule(lr){2-7} \cmidrule(lr){8-13}  
  & \multicolumn{2}{c}{PredCLS} & \multicolumn{2}{c}{SGCLS} & \multicolumn{2}{c}{SGDET} & \multicolumn{2}{c}{PredCLS} & \multicolumn{2}{c}{SGCLS} & \multicolumn{2}{c}{SGDET} \cr
    \cmidrule(lr){2-3} \cmidrule(lr){4-5} \cmidrule(lr){6-7} \cmidrule(lr){8-9} \cmidrule(lr){10-11} \cmidrule(lr){12-13} 
  &mR@10  &mR@50 &mR@10  &mR@50 &mR@10 &mR@50 &mR@10 &mR@50 &mR@10  &mR@50 &mR@10  &mR@50\\
  \hline \hline

  RelDN~\shortcite{zhang2019graphical} &6.2 &6.2 &3.4 &3.4 &3.3 &3.3 &31.2 &75.5 &18.6 &42.6 &7.5 &37.7 \\

  TRACE~\shortcite{teng2021target} &15.2 &15.2 &8.9 &8.9 &8.2 &8.2 &50.9  &82.7 &31.9 &46.3 &22.8 &41.8\\
  STTran~\shortcite{cong2021spatial}&37.8 &40.2 &27.2  &28.0 &16.6  &22.2 &51.4  &82.7 &40.7 &58.8 &20.9  &39.2 \\
  TPI~\shortcite{wang2022dynamic}&37.3 &40.6 &28.3 &29.3 &15.6&21.8 &- &- &- & -&- & -\\

  TEMP~\shortcite{nag2023unbiased} &42.9  &46.3 &34.0 &35.2 &18.5&23.7 &61.5  &98.0 &48.3  &66.4 &24.7  &43.7\\

  \hline
   {\bf TD$^2$-Net (P)} &41.9 &44.8 &33.9 &34.9  &17.2 &22.3 &61.0 &96.4 &50.1&67.9&23.2&42.1\\
   {\bf TD$^2$-Net} &\bf 54.2  &\bf 57.1 &\bf 40.9 &\bf 42.0  &\bf 20.4&\bf 26.1&\bf 68.3 &\bf 98.2 &\bf 51.4 &\bf 69.1&\bf 27.9 &\bf 46.3\\
  
\midrule[1pt]
\bottomrule[2pt]
\end{tabular}
\caption{ Comparison on the mR@K metric between various methods across all the 26 relationship categories. Note that we adopt the same evaluation metric as TEMP~\cite{nag2023unbiased}. TD$^2$-Net (P) indicate that we set $\omega_{cb}$ as 1 in AR-Loss. The best methods under each setting are marked according to formats.}
\label{tb2}
\end{table*}

\noindent\textbf{Implementation Details.} To ensure compatibility with previous state-of-the-art architectures, we follow STTran~\cite{cong2021spatial} and use Faster R-CNN~\cite{ren2015faster} based on ResNet-101~\cite{he2016deep} as the backbone for object detection. During training, we utilize the AdamW optimizer~\cite{loshchilov2017decoupled} with an initial learning rate of $1e^{-5}$ and a batch size of $1$. The model is trained for $10$ epochs. Additionally, we apply gradient clipping, restricting the gradients to a maximum norm of $5$. In the Eq.~\eqref{ntrans} and Eq.~\eqref{ntrans_s}, we set parameters $M$ = $N$ = 3.

\subsection{Comparisons with State-of-the-Art Methods}\label{sota}
Table~\ref{tb1} shows that TD$^2$-Net (P) outperforms all state-of-the-art methods on various metrics. Specifically, TD$^2$-Net (P) outperforms the best ImgSGG method, named GPS-Net, by 3.3 $\%$, 5.8$\%$, and 4.0$\%$ at R@10 on PRECLS, SGCLS, and SGDET, respectively. Moreover, even when compared with the best unbiased dynamic SGG method TEMP, TD$^2$-Net (P) still demonstrates a performance improvement of 3.9$\%$ at R@10 on SGCLS task. 

\begin{figure}[t]
\centering
\includegraphics[width=1.\linewidth]{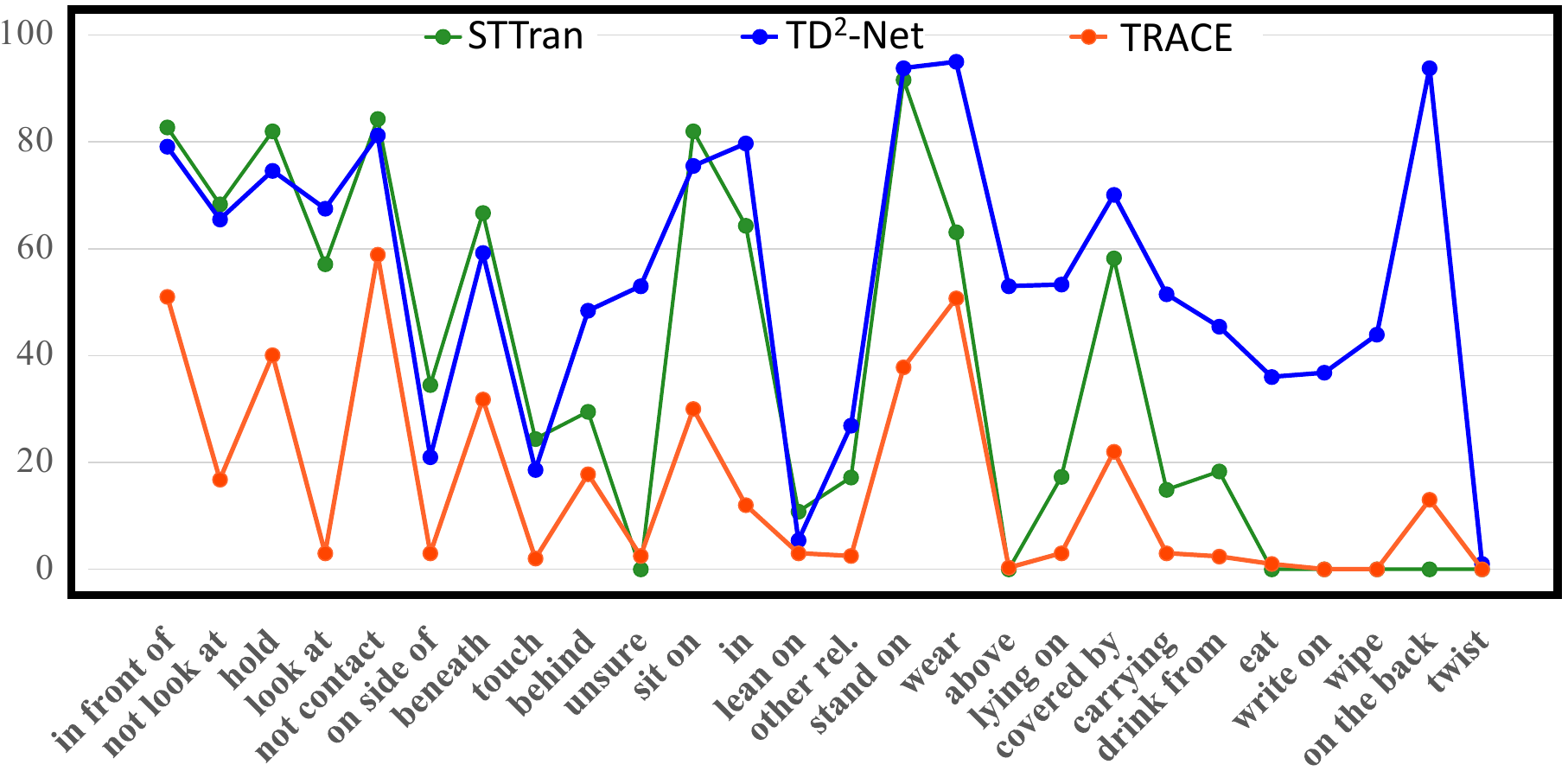}
\caption{Comparative per class performance for PREDCLS task. Results are in terms of R@10 under With Constraint.}
\label{fig:qual-1}
\end{figure}

Due to the class imbalance problem in Action Genome, previous works usually achieve low performance for less frequent categories. Hence, we conduct an experiment utilizing the mR@K as evaluation metric~\cite{nag2023unbiased} for all three SGG tasks under both {\bf{With Constraint}} and {\bf{No Constraints}} settings. We also relied on email communications with the authors of several papers on the mR values where the source code are not publicly available. As shown in Table~\ref{tb2}, TD$^2$-Net shows a large absolute gain for the Mean Recall metric, which indicates that TD$^2$-Net has advantages in handling the class imbalance problem of dynamic SGG. In more detail, TD$^2$-Net outperform one very recent unbiased dynamic SGG method, named TEMP~\cite{nag2023unbiased}, by 12.7$\%$ at mR@10 on PREDCLS under with constraint setting. To illustrate this advantage more vividly, we present the R@10 improvement of each predicate category compared with STTran~\cite{cong2021spatial}, and TRACE~\cite{teng2021target} for PREDCLS under {\bf{With Constraints}} setting in Figure~\ref{fig:qual-1}. These improvements are much larger for minority relationship categories. We owe this advantage to the power of the AR-Loss. Overall, TD$^2$-Net does not compromise Recall values and achieves comparable or better performance than the existing methods, which aim to achieve high Recall values without considering label bias issues.

\begin{figure*}[ht]
\begin{center}
    \centering
    \includegraphics[scale=0.53]{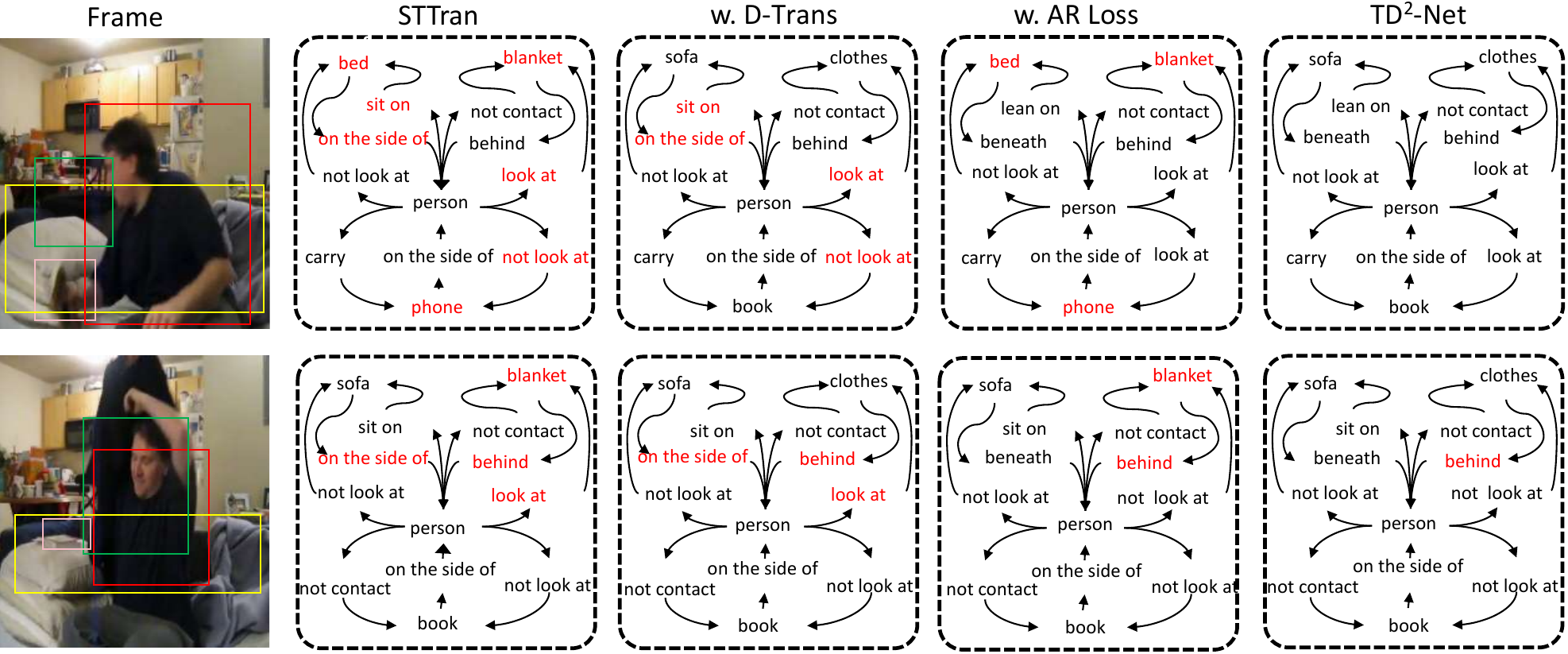}
    \captionof{figure}{Qualitative comparisons between TD$^2$-Net and STTran~\cite{cong2021spatial}. Specifically, we show the comparisons at R@100 in the SGCLS setting. The black color indicates correctly classified objects or predicates; the red indicates those that have been misclassified. Best viewed in color.}\label{qual-2}
\end{center}
\vspace{-15pt}
\end{figure*}

\subsection{Ablation Studies}\label{ablation_all}
We conduct four ablation studies to verify the effectiveness of our proposed methods. The results of the ablation studies are summarized in four tables: Table~\ref{tb-q1}, Table~\ref{tb-q2}, Table~\ref{tb-qq}, and Table~\ref{tb-q3}. It is worth noting that we have adopted TD$^2$-Net (P), which sets $\omega_{cb}$ as 1 in AR-Loss, in Table~\ref{tb-q2} and Table~\ref{tb-qq} to demonstrate the significant of each component of D-Trans.

\begin{table}[t]
\small
\centering
\setlength{\tabcolsep}{1.65mm}
\renewcommand{\arraystretch}{1.1}\begin{tabular}
{l|cc|cc|cc}
\toprule[2pt]
\midrule[1pt]
\multicolumn{1}{l|}{} &\multicolumn{2}{c|}{Module} & \multicolumn{2}{c|}{SGCLS} & \multicolumn{2}{c}{PredCLS} \\
 \multicolumn{1}{c|}{Exp} &\multicolumn{1}{c}{D-Trans} & \multicolumn{1}{c|}{AR}  & R@10 &  \multicolumn{1}{c|}{mR@10}  & \multicolumn{1}{l}{R@10} &   \multicolumn{1}{l}{mR@10} \\ \hline\hline

\multicolumn{1}{c|}{1}  & \multicolumn{1}{c}{-} & \multicolumn{1}{c|}{-} &46.4 & \multicolumn{1}{c|}{27.2} &\multicolumn{1}{c}{{68.6}}  &\multicolumn{1}{c}{{37.8}}\\

\multicolumn{1}{c|}{2}  & \multicolumn{1}{c}{\checkmark} & \multicolumn{1}{c|}{-} &49.7 & \multicolumn{1}{c|}{29.5} &\multicolumn{1}{c}{{68.9}}  &\multicolumn{1}{c}{{40.4}}\\

\multicolumn{1}{c|}{3}  & \multicolumn{1}{c}{-} & \multicolumn{1}{c|}{\semichecked} &47.3 & \multicolumn{1}{c|}{30.6} &\multicolumn{1}{c}{{69.7}}  &\multicolumn{1}{c}{{40.3}}\\

\multicolumn{1}{c|}{4}  & \multicolumn{1}{c}{-} & \multicolumn{1}{c|}{\checkmark} &45.7 & \multicolumn{1}{c|}{37.3} &\multicolumn{1}{c}{{67.3}}  &\multicolumn{1}{c}{{53.1}}  \\
\hline
\multicolumn{1}{c|}{5}  & \multicolumn{1}{c}{\checkmark} & \multicolumn{1}{c|}{\semichecked} &{\bf 51.1} & \multicolumn{1}{c|}{33.9} &\multicolumn{1}{c}{{\bf 70.1}}  &\multicolumn{1}{c}{{42.2}}\\

\multicolumn{1}{c|}{6}  & \multicolumn{1}{c}{\checkmark} & \multicolumn{1}{c|}{\checkmark} &49.1 & \multicolumn{1}{c|}{\bf 40.9} &\multicolumn{1}{c}{{67.8}}  &\multicolumn{1}{c}{{\bf 54.2}}\\

\midrule[1pt]
\bottomrule[2pt]
\end{tabular}
\caption{Ablation studies. We consistently adopt the same object detection backbone as in \cite{cong2021spatial}. ``\semichecked" denotes that we set $\omega_{cb}$ as 1 in AR-Loss.}
\label{tb-q1}
\end{table}

\noindent\textbf{Effectiveness of the Proposed Modules.} We first perform an ablation study to justify the effectiveness of D-Trans and AR-Loss. The results are summarized in Table~\ref{tb-q1}. Exp 1 in Table~\ref{tb-q1} shows the baseline performance based on STTran~\cite{cong2021spatial}. To facilitate fair comparison, all the other settings remain the same as TD$^2$-Net. Exps 2-4 show that each module helps promote dynamic SGG performance. The best performance is achieved when both modules are involved. Note that D-Trans and AR-Loss are primarily designed to refine object and relationship representations, respectively. Therefore, D-Trans helps the model achieve outstanding SGCLS performance, which heavily depends on the object classification ability. Meanwhile, AR-Loss enables the model to achieve a significant performance gain on the PREDCLS task, mainly relying on relationship prediction power.

\begin{table}[t]
\small
\centering
\setlength{\tabcolsep}{2.0mm}
\renewcommand{\arraystretch}{1.1}\begin{tabular}
{ccc|clclclclcl}
\toprule[2pt]\midrule[1pt]
\multicolumn{3}{c|}{} & \multicolumn{2}{c}{$K$} & \multicolumn{2}{c}{4} & \multicolumn{2}{c}{6} & \multicolumn{2}{c}{8}  & \multicolumn{2}{c}{10}\\ \hline\hline
\multicolumn{1}{l}{} &  &  & \multicolumn{2}{c}{R@10} & \multicolumn{2}{c}{49.6} & \multicolumn{2}{c}{50.4} & \multicolumn{2}{c}{\bf51.1}& \multicolumn{2}{c}{50.9} \\
\multicolumn{3}{c|}{SGCLS} & \multicolumn{2}{c}{R@20} & \multicolumn{2}{c}{50.7} & \multicolumn{2}{c}{51.4} & \multicolumn{2}{c}{\bf52.1}& \multicolumn{2}{c}{51.9}\\
\multicolumn{1}{l}{} & \multicolumn{1}{l}{} & \multicolumn{1}{l|}{} & \multicolumn{2}{c}{R@50} & \multicolumn{2}{c}{50.7} & \multicolumn{2}{c}{51.4} & \multicolumn{2}{c}{\bf52.1}& \multicolumn{2}{c}{51.9} \\ \midrule[1pt]
\bottomrule[2pt]
\end{tabular}
\caption{Evaluation on the value of Top-K in Eq.~\eqref{selector}.}
\label{tb-q2}
\end{table}

\begin{table}[t]
\small
\centering
\setlength{\tabcolsep}{1.2mm}
\renewcommand{\arraystretch}{1.1}\begin{tabular}
{ccc|clclclclcl}
\toprule[2pt]\midrule[1pt]
\multicolumn{3}{c|}{} & \multicolumn{2}{c}{} & \multicolumn{2}{c}{w/o D-Trans}& \multicolumn{2}{c}{Eq.~\eqref{score_1}} & \multicolumn{2}{c}{Eq.~\eqref{selector}}& \multicolumn{2}{c}{Full}\\ \hline\hline
 \multicolumn{1}{l}{} &  &  & \multicolumn{2}{c}{R@10} & \multicolumn{2}{c}{47.3}& \multicolumn{2}{c}{49.1} & \multicolumn{2}{c}{50.7}  & \multicolumn{2}{c}{\bf 51.1}\\
\multicolumn{3}{c|}{SGCLS} & \multicolumn{2}{c}{R@20} & \multicolumn{2}{c}{48.4}& \multicolumn{2}{c}{50.2} & \multicolumn{2}{c}{51.7} & \multicolumn{2}{c}{\bf 52.1}\\
\multicolumn{1}{l}{} & \multicolumn{1}{l}{} & \multicolumn{1}{l|}{} & \multicolumn{2}{c}{R@50}& \multicolumn{2}{c}{48.4} & \multicolumn{2}{c}{50.2} & \multicolumn{2}{c}{51.7}& \multicolumn{2}{c}{\bf 52.1}\\ \midrule[1pt]
\bottomrule[2pt]
\end{tabular}
\caption{Design Choices for the D-Trans module.}
\label{tb-qq}
\end{table}

\begin{table}[t]
\small
\centering
\setlength{\tabcolsep}{1.45mm}
\renewcommand{\arraystretch}{1.1}\begin{tabular}
{ccc|cccccccccccc}
\toprule[2pt]
\midrule[1pt]
\multicolumn{3}{c|}{} & \multicolumn{2}{c}{} & \multicolumn{2}{c}{Focal} & \multicolumn{2}{c}{BCE} & \multicolumn{2}{c}{MLM} & \multicolumn{2}{c}{$\omega_{cb}=1$}& \multicolumn{2}{c}{AR} \\ \hline\hline
 &  &  & \multicolumn{2}{c}{R@10} & \multicolumn{2}{c}{69.4} & \multicolumn{2}{c}{69.2} & \multicolumn{2}{c}{69.0} & \multicolumn{2}{c}{\bf70.1} & \multicolumn{2}{c}{67.8}  \\
\multicolumn{3}{c|}{} & \multicolumn{2}{c}{R@20} & \multicolumn{2}{c}{72.4}& \multicolumn{2}{c}{72.1} & \multicolumn{2}{c}{71.9} & \multicolumn{2}{c}{\bf73.1}  & \multicolumn{2}{c}{70.8} \\
\multicolumn{3}{c|}{} & \multicolumn{2}{c}{R@50} & \multicolumn{2}{c}{72.4}  & \multicolumn{2}{c}{72.1}& \multicolumn{2}{c}{71.9} & \multicolumn{2}{c}{\bf73.1}  & \multicolumn{2}{c}{70.8} \\\cline{4-15}
\multicolumn{3}{c|}{\multirow{0.1}{*}{PREDCLS}} & \multicolumn{2}{c}{mR@10} & \multicolumn{2}{c}{42.1} & \multicolumn{2}{c}{41.2} & \multicolumn{2}{c}{40.4} & \multicolumn{2}{c}{41.9} & \multicolumn{2}{c}{\bf54.2} \\
\multicolumn{1}{l}{} &  &  & \multicolumn{2}{c}{mR@20} & \multicolumn{2}{c}{44.3}& \multicolumn{2}{c}{43.4} & \multicolumn{2}{c}{42.7} & \multicolumn{2}{c}{44.8}  & \multicolumn{2}{c}{\bf57.1}\\
\multicolumn{1}{l}{} & \multicolumn{1}{l}{} & \multicolumn{1}{l|}{} & \multicolumn{2}{c}{mR@50} & \multicolumn{2}{c}{44.3}& \multicolumn{2}{c}{43.4} & \multicolumn{2}{c}{42.7} & \multicolumn{2}{c}{44.8}& \multicolumn{2}{c}{\bf57.1}  \\

\midrule[1pt]
\bottomrule[2pt]
\end{tabular}
\caption{Evaluation on different choices of loss function.}
\label{tb-q3}
\vspace{-15pt}
\end{table}

\noindent\textbf{Evaluation on Hyperparameters for D-Trans.}
we verify the impact of the hyperparameters of the D-Trans modules. As shown in Table~\ref{tb-q2}, TD$^2$-Net (P) achieves the best performance when K is set to 8 in the differentiable Top-$\rm K$ frame selector. However, if this threshold is exceeded, the model's memory usage increases with decreased benefits. More details can be found in the supplemental file.

\noindent\textbf{Design Choices for the D-Trans Module.}
In Table~\ref{tb-qq}, we compare the performance of D-Trans with and without the two object-matching strategies described in Eq.~\eqref{score_1} and Eq.~\eqref{selector}. ``w/o D-Trans" denotes that we remove the D-Trans module in TD$^2$-Net (P). ``Full" represents that we simultaneously utilize Eq.~\eqref{score_1} and Eq.~\eqref{selector} in differentiable Top-K object selector for TD$^2$-Net (P). Experimental results in Table~\ref{tb-qq} show that the two object-matching strategies consistently achieve better performance. Therefore, the effectiveness of differentiable Top-K object selector is justified.

\noindent\textbf{Comparisons between Four Loss Functions.}
We compare the performance of the Focal-Loss~\cite{lin2017focal}, AR-Loss ($\omega_{cb}$=1), AR-Loss, and the other two kinds of loss functions which are widely utilized in the existing VidSGG methods: MLM-Loss~\cite{cong2021spatial}, BCE-Loss~\cite{nag2023unbiased}. As shown in Table~\ref{tb-q3}, AR-Loss ($\omega_{cb}$=1) achieves the best performance in terms of R@$K$, as it effectively mitigates the issue of positive-negative imbalance. Additionally, when we focus on mR@$K$, AR-Loss outperforms the other methods. This can be attributed to the fact that AR-Loss effectively mitigates the issue of label bias.

\noindent\textbf{Qualitative Evaluation.} 
Figure~\ref{qual-2} presents the qualitative results for the dynamic scene graph generation. The five columns from left to right are RGB frame, scene graph generated by STTran, STTran with D-Trans, STTran with AR-Loss, and scene graph generated by TD$^2$-Net, respectively. As can be seen from the third column of Figure~\ref{qual-2}, STTran with D-Trans produces superior object predictions compared to STTran for items such as ``sofa'', ``book'', and ``clothes'' that are challenging to identify from their proposals. Therefore, we owe this performance gain to the D-Trans module that utilizes robust context information to enhance the object's representation. In the fourth column of Figure~\ref{qual-2}, the improvements in predicates predictions, such as ``lean on'' and ``behind'', can be attributed to the contributions of AR-Loss. Furthermore, as illustrated in the rightmost column, TD$^2$-Net demonstrates superior overall performance.

\subsection{Conclusion}\label{conclusion}
In this paper, we propose a new model called TD$^2$-Net, which is designed to handle two critical issues in dynamic SGG: contextual noise and label bias. To address the contextual noise issue, we introduce a D-Trans module that uses a differentiable Top-K object selector to choose the most relevant neighborhood for each object. Then, we could enhance object representation with robust contextual information via spatio-temporal message passing. To mitigate the head-tail and positive-negative imbalance in relationship prediction, we introduce an AR-Loss which incorporates asymmetry focusing factors and sample volume to adjust the sample weights. Through extensive experiments on the Action Genome dataset, we demonstrate the effectiveness of our approach.

\section{Acknowledgments}
This work is supported by the Guangzhou basic and applied basic research scheme (No: 2024A04J3367), and the NSF of China (No: 62002090).

\bibliography{aaai24}

\begin{thebibliography}{37}
\providecommand{\natexlab}[1]{#1}

\bibitem[{Buch et~al.(2022)Buch, Eyzaguirre, Gaidon, Wu, Fei-Fei, and Niebles}]{buch2022revisiting}
Buch, S.; Eyzaguirre, C.; Gaidon, A.; Wu, J.; Fei-Fei, L.; and Niebles, J.~C. 2022.
\newblock Revisiting the" video" in video-language understanding.
\newblock In \emph{Proceedings of the IEEE/CVF Conference on Computer Vision and Pattern Recognition}, 2917--2927.

\bibitem[{Cong et~al.(2021)Cong, Liao, Ackermann, Rosenhahn, and Yang}]{cong2021spatial}
Cong, Y.; Liao, W.; Ackermann, H.; Rosenhahn, B.; and Yang, M.~Y. 2021.
\newblock Spatial-temporal transformer for dynamic scene graph generation.
\newblock In \emph{Proceedings of the IEEE/CVF international conference on computer vision}, 16372--16382.

\bibitem[{Cui et~al.(2019)Cui, Jia, Lin, Song, and Belongie}]{cui2019class}
Cui, Y.; Jia, M.; Lin, T.-Y.; Song, Y.; and Belongie, S. 2019.
\newblock Class-balanced loss based on effective number of samples.
\newblock In \emph{Proceedings of the IEEE/CVF conference on computer vision and pattern recognition}, 9268--9277.

\bibitem[{Feng et~al.(2023)Feng, Mostafa, Nassar, Majumdar, and Tripathi}]{feng2023exploiting}
Feng, S.; Mostafa, H.; Nassar, M.; Majumdar, S.; and Tripathi, S. 2023.
\newblock Exploiting long-term dependencies for generating dynamic scene graphs.
\newblock In \emph{Proceedings of the IEEE/CVF Winter Conference on Applications of Computer Vision}, 5130--5139.

\bibitem[{Gao et~al.(2023{\natexlab{a}})Gao, Zhou, Ji, Zhu, Yang, and Shou}]{gao2023mist}
Gao, D.; Zhou, L.; Ji, L.; Zhu, L.; Yang, Y.; and Shou, M.~Z. 2023{\natexlab{a}}.
\newblock MIST: Multi-modal Iterative Spatial-Temporal Transformer for Long-form Video Question Answering.
\newblock In \emph{Proceedings of the IEEE/CVF Conference on Computer Vision and Pattern Recognition}, 14773--14783.

\bibitem[{Gao et~al.(2021)Gao, Chen, Huang, and Xiao}]{gao2021video}
Gao, K.; Chen, L.; Huang, Y.; and Xiao, J. 2021.
\newblock Video relation detection via tracklet based visual transformer.
\newblock In \emph{Proceedings of the 29th ACM international conference on multimedia}, 4833--4837.

\bibitem[{Gao et~al.(2022)Gao, Chen, Niu, Shao, and Xiao}]{gao2022classification}
Gao, K.; Chen, L.; Niu, Y.; Shao, J.; and Xiao, J. 2022.
\newblock Classification-then-grounding: Reformulating video scene graphs as temporal bipartite graphs.
\newblock In \emph{Proceedings of the IEEE/CVF Conference on Computer Vision and Pattern Recognition}, 19497--19506.

\bibitem[{Gao et~al.(2023{\natexlab{b}})Gao, Chen, Zhang, Xiao, and Sun}]{gao2023compositional}
Gao, K.; Chen, L.; Zhang, H.; Xiao, J.; and Sun, Q. 2023{\natexlab{b}}.
\newblock Compositional prompt tuning with motion cues for open-vocabulary video relation detection.
\newblock \emph{arXiv preprint arXiv:2302.00268}.

\bibitem[{He et~al.(2016)He, Zhang, Ren, and Sun}]{he2016deep}
He, K.; Zhang, X.; Ren, S.; and Sun, J. 2016.
\newblock Deep residual learning for image recognition.
\newblock In \emph{Proceedings of the IEEE conference on computer vision and pattern recognition}, 770--778.

\bibitem[{Jang, Gu, and Poole(2016)}]{jang2016categorical}
Jang, E.; Gu, S.; and Poole, B. 2016.
\newblock Categorical reparameterization with gumbel-softmax.
\newblock \emph{arXiv preprint arXiv:1611.01144}.

\bibitem[{Ji et~al.(2020)Ji, Krishna, Fei-Fei, and Niebles}]{ji2020action}
Ji, J.; Krishna, R.; Fei-Fei, L.; and Niebles, J.~C. 2020.
\newblock Action genome: Actions as compositions of spatio-temporal scene graphs.
\newblock In \emph{Proceedings of the IEEE/CVF Conference on Computer Vision and Pattern Recognition}, 10236--10247.

\bibitem[{Johnson et~al.(2015)Johnson, Krishna, Stark, Li, Shamma, Bernstein, and Fei-Fei}]{johnson2015image}
Johnson, J.; Krishna, R.; Stark, M.; Li, L.-J.; Shamma, D.; Bernstein, M.; and Fei-Fei, L. 2015.
\newblock Image retrieval using scene graphs.
\newblock In \emph{Proceedings of the IEEE conference on computer vision and pattern recognition}, 3668--3678.

\bibitem[{Li et~al.(2021)Li, Zhang, Wan, and He}]{li2021bipartite}
Li, R.; Zhang, S.; Wan, B.; and He, X. 2021.
\newblock Bipartite graph network with adaptive message passing for unbiased scene graph generation.
\newblock In \emph{Proceedings of the IEEE/CVF Conference on Computer Vision and Pattern Recognition}, 11109--11119.

\bibitem[{Li et~al.(2017)Li, Ouyang, Zhou, Wang, and Wang}]{li2017scene}
Li, Y.; Ouyang, W.; Zhou, B.; Wang, K.; and Wang, X. 2017.
\newblock Scene graph generation from objects, phrases and region captions.
\newblock In \emph{Proceedings of the IEEE international conference on computer vision}, 1261--1270.

\bibitem[{Li, Yang, and Xu(2022)}]{li2022dynamic}
Li, Y.; Yang, X.; and Xu, C. 2022.
\newblock Dynamic scene graph generation via anticipatory pre-training.
\newblock In \emph{Proceedings of the IEEE/CVF Conference on Computer Vision and Pattern Recognition}, 13874--13883.

\bibitem[{Lin et~al.(2017)Lin, Goyal, Girshick, He, and Doll{\'a}r}]{lin2017focal}
Lin, T.-Y.; Goyal, P.; Girshick, R.; He, K.; and Doll{\'a}r, P. 2017.
\newblock Focal loss for dense object detection.
\newblock In \emph{Proceedings of the IEEE international conference on computer vision}, 2980--2988.

\bibitem[{Lin et~al.(2020)Lin, Ding, Zeng, and Tao}]{lin2020gps}
Lin, X.; Ding, C.; Zeng, J.; and Tao, D. 2020.
\newblock Gps-net: Graph property sensing network for scene graph generation.
\newblock In \emph{Proceedings of the IEEE/CVF Conference on Computer Vision and Pattern Recognition}, 3746--3753.

\bibitem[{Lin et~al.(2022{\natexlab{a}})Lin, Ding, Zhan, Li, and Tao}]{lin2022hl}
Lin, X.; Ding, C.; Zhan, Y.; Li, Z.; and Tao, D. 2022{\natexlab{a}}.
\newblock Hl-net: Heterophily learning network for scene graph generation.
\newblock In \emph{proceedings of the IEEE/CVF conference on computer vision and pattern recognition}, 19476--19485.

\bibitem[{Lin et~al.(2022{\natexlab{b}})Lin, Ding, Zhang, Zhan, and Tao}]{lin2022ru}
Lin, X.; Ding, C.; Zhang, J.; Zhan, Y.; and Tao, D. 2022{\natexlab{b}}.
\newblock Ru-net: Regularized unrolling network for scene graph generation.
\newblock In \emph{Proceedings of the IEEE/CVF Conference on Computer Vision and Pattern Recognition}, 19457--19466.

\bibitem[{Liu et~al.(2020)Liu, Jin, Xu, Gong, and Mu}]{liu2020beyond}
Liu, C.; Jin, Y.; Xu, K.; Gong, G.; and Mu, Y. 2020.
\newblock Beyond short-term snippet: Video relation detection with spatio-temporal global context.
\newblock In \emph{Proceedings of the IEEE/CVF conference on computer vision and pattern recognition}, 10840--10849.

\bibitem[{Loshchilov and Hutter(2017)}]{loshchilov2017decoupled}
Loshchilov, I.; and Hutter, F. 2017.
\newblock Decoupled weight decay regularization.
\newblock \emph{arXiv preprint arXiv:1711.05101}.

\bibitem[{Lu et~al.(2016)Lu, Krishna, Bernstein, and Fei-Fei}]{lu2016visual}
Lu, C.; Krishna, R.; Bernstein, M.; and Fei-Fei, L. 2016.
\newblock Visual relationship detection with language priors.
\newblock In \emph{Computer Vision--ECCV 2016: 14th European Conference, Amsterdam, The Netherlands, October 11--14, 2016, Proceedings, Part I 14}, 852--869. Springer.

\bibitem[{Nag et~al.(2023)Nag, Min, Tripathi, and Roy-Chowdhury}]{nag2023unbiased}
Nag, S.; Min, K.; Tripathi, S.; and Roy-Chowdhury, A.~K. 2023.
\newblock Unbiased Scene Graph Generation in Videos.
\newblock In \emph{Proceedings of the IEEE/CVF Conference on Computer Vision and Pattern Recognition}, 22803--22813.

\bibitem[{Ren et~al.(2015)Ren, He, Girshick, and Sun}]{ren2015faster}
Ren, S.; He, K.; Girshick, R.; and Sun, J. 2015.
\newblock Faster r-cnn: Towards real-time object detection with region proposal networks.
\newblock \emph{Advances in neural information processing systems}, 28.

\bibitem[{Shang et~al.(2019)Shang, Di, Xiao, Cao, Yang, and Chua}]{shang2019annotating}
Shang, X.; Di, D.; Xiao, J.; Cao, Y.; Yang, X.; and Chua, T.-S. 2019.
\newblock Annotating objects and relations in user-generated videos.
\newblock In \emph{Proceedings of the 2019 on International Conference on Multimedia Retrieval}, 279--287.

\bibitem[{Shang et~al.(2017)Shang, Ren, Guo, Zhang, and Chua}]{shang2017video}
Shang, X.; Ren, T.; Guo, J.; Zhang, H.; and Chua, T.-S. 2017.
\newblock Video visual relation detection.
\newblock In \emph{Proceedings of the 25th ACM international conference on Multimedia}, 1300--1308.

\bibitem[{Sigurdsson et~al.(2016)Sigurdsson, Varol, Wang, Farhadi, Laptev, and Gupta}]{sigurdsson2016hollywood}
Sigurdsson, G.~A.; Varol, G.; Wang, X.; Farhadi, A.; Laptev, I.; and Gupta, A. 2016.
\newblock Hollywood in homes: Crowdsourcing data collection for activity understanding.
\newblock In \emph{Computer Vision--ECCV 2016: 14th European Conference, Amsterdam, The Netherlands, October 11--14, 2016, Proceedings, Part I 14}, 510--526. Springer.

\bibitem[{Tang et~al.(2019)Tang, Zhang, Wu, Luo, and Liu}]{tang2019learning}
Tang, K.; Zhang, H.; Wu, B.; Luo, W.; and Liu, W. 2019.
\newblock Learning to compose dynamic tree structures for visual contexts.
\newblock In \emph{Proceedings of the IEEE/CVF conference on computer vision and pattern recognition}, 6619--6628.

\bibitem[{Teng et~al.(2021)Teng, Wang, Li, and Wu}]{teng2021target}
Teng, Y.; Wang, L.; Li, Z.; and Wu, G. 2021.
\newblock Target adaptive context aggregation for video scene graph generation.
\newblock In \emph{Proceedings of the IEEE/CVF International Conference on Computer Vision}, 13688--13697.

\bibitem[{Vaswani et~al.(2017)Vaswani, Shazeer, Parmar, Uszkoreit, Jones, Gomez, Kaiser, and Polosukhin}]{vaswani2017attention}
Vaswani, A.; Shazeer, N.; Parmar, N.; Uszkoreit, J.; Jones, L.; Gomez, A.~N.; Kaiser, {\L}.; and Polosukhin, I. 2017.
\newblock Attention is all you need.
\newblock \emph{Advances in neural information processing systems}, 30.

\bibitem[{Wang et~al.(2022)Wang, Gao, Lyu, Guo, Zeng, and Song}]{wang2022dynamic}
Wang, S.; Gao, L.; Lyu, X.; Guo, Y.; Zeng, P.; and Song, J. 2022.
\newblock Dynamic scene graph generation via temporal prior inference.
\newblock In \emph{Proceedings of the 30th ACM International Conference on Multimedia}, 5793--5801.

\bibitem[{Xiao et~al.(2022)Xiao, Zhou, Chua, and Yan}]{xiao2022video}
Xiao, J.; Zhou, P.; Chua, T.-S.; and Yan, S. 2022.
\newblock Video graph transformer for video question answering.
\newblock In \emph{European Conference on Computer Vision}, 39--58. Springer.

\bibitem[{Xie et~al.(2020)Xie, Dai, Chen, Dai, Zhao, Zha, Wei, and Pfister}]{xie2020differentiable}
Xie, Y.; Dai, H.; Chen, M.; Dai, B.; Zhao, T.; Zha, H.; Wei, W.; and Pfister, T. 2020.
\newblock Differentiable top-k with optimal transport.
\newblock \emph{Advances in Neural Information Processing Systems}, 33: 20520--20531.

\bibitem[{Yang et~al.(2020)Yang, Gao, Zhang, and Cai}]{yang2020hierarchical}
Yang, X.; Gao, C.; Zhang, H.; and Cai, J. 2020.
\newblock Hierarchical scene graph encoder-decoder for image paragraph captioning.
\newblock In \emph{Proceedings of the 28th ACM International Conference on Multimedia}, 4181--4189.

\bibitem[{Zellers et~al.(2018)Zellers, Yatskar, Thomson, and Choi}]{zellers2018neural}
Zellers, R.; Yatskar, M.; Thomson, S.; and Choi, Y. 2018.
\newblock Neural motifs: Scene graph parsing with global context.
\newblock In \emph{Proceedings of the IEEE conference on computer vision and pattern recognition}, 5831--5840.

\bibitem[{Zhang et~al.(2019)Zhang, Shih, Elgammal, Tao, and Catanzaro}]{zhang2019graphical}
Zhang, J.; Shih, K.~J.; Elgammal, A.; Tao, A.; and Catanzaro, B. 2019.
\newblock Graphical contrastive losses for scene graph parsing.
\newblock In \emph{Proceedings of the IEEE/CVF Conference on Computer Vision and Pattern Recognition}, 11535--11543.

\bibitem[{Zheng et~al.(2023)Zheng, Lyu, Gao, Dai, and Song}]{zheng2023prototype}
Zheng, C.; Lyu, X.; Gao, L.; Dai, B.; and Song, J. 2023.
\newblock Prototype-based Embedding Network for Scene Graph Generation.
\newblock In \emph{Proceedings of the IEEE/CVF Conference on Computer Vision and Pattern Recognition}, 22783--22792.

\end{thebibliography}

\end{document}